\documentclass[10pt,twocolumn,letterpaper]{article}

\usepackage{iccv}
\usepackage{times}
\usepackage{epsfig}
\usepackage{graphicx}
\usepackage{amsmath}
\usepackage{amssymb}
\usepackage{times}
\usepackage{epsfig}
\usepackage{graphicx}
\usepackage{amsmath}
\usepackage{amssymb}
\usepackage{subcaption}
\usepackage[table]{xcolor} 

\usepackage[breaklinks=true,bookmarks=false,colorlinks = true]{hyperref}

\iccvfinalcopy % *** Uncomment this line for the final submission

\makeatletter
\newcommand*\titleheader[1]{\gdef\@titleheader{#1}}
\AtBeginDocument{%f
  \let\st@red@title\@title
  \def\@title{%
    \bgroup\normalfont\large\centering\@titleheader\par\egroup
    \vskip1.5em\st@red@title}
}
\makeatother

\title{ Efficient Convolutional Network Learning using \\ Parametric Log based Dual-Tree Wavelet ScatterNet}
\titleheader{To Appear in the IEEE International Conference on Computer Vision Workshops (ICCVW) 2017}
\author{maxmin}

\begin{document}

\author{Amarjot Singh, Nick Kingsbury\\
Signal Processing Group, Department of Engineering, University of Cambridge, U.K.
\\
{\tt\small as2436@cam.ac.uk, ngk10@cam.ac.uk}}
\maketitle

\begin{abstract}
We propose a DTCWT ScatterNet Convolutional Neural Network (DTSCNN) formed by replacing the first few layers of a CNN network with a parametric log based DTCWT ScatterNet. The ScatterNet extracts edge based invariant representations that are used by the later layers of the CNN to learn high-level features. This improves the training of the network as the later layers can learn more complex patterns from the start of learning because the edge representations are already present. The efficient learning of the DTSCNN network is demonstrated on CIFAR-10 and Caltech-101 datasets. The generic nature of the ScatterNet front-end is shown by an equivalent performance to pre-trained CNN front-ends. A comparison with the state-of-the-art on CIFAR-10 and Caltech-101 datasets is also presented.
\end{abstract}

%%%%%%%%% BODY TEXT
\section{Introduction}
Deep Convolutional Neural Networks (DCNNs) have made great advances at numerous classification~\cite{Krizhevsky} and regression~\cite{regression} tasks in computer vision and speech applications over the past few years. However, these models produce state-of-the-art results only for large datasets and tend to overfit~\cite{overfit} on many other applications such as the analysis of hyperspectral images~\cite{satellite}, stock market prediction~\cite{su}, medical data analysis~\cite{medical} etc due to the small training datasets. 

Two primary approaches have been utilized to train DCNNs effectively for applications with small training datasets: (i) Data augmentation and synthetic data generation, and (ii) Transfer Learning. Training CNNs on synthetic datasets may not learn potentially useful patterns of real data as often the feature distribution of synthetic data generated shifts away from the real data~\cite{syn}. On the other hand, transfer Learning aims to extract the knowledge from one or more source tasks and applies the knowledge to a target task. The weights of the CNN are initialized with those from a network trained for related tasks before fine-tuning them using the target dataset~\cite{finetune}. These Networks have resulted in excellent embeddings, which generalize well to new categories~\cite{embed}.

\begin{figure*}[t!] 
\centering    
\includegraphics[width = 17 cm, height = 9.5 cm]{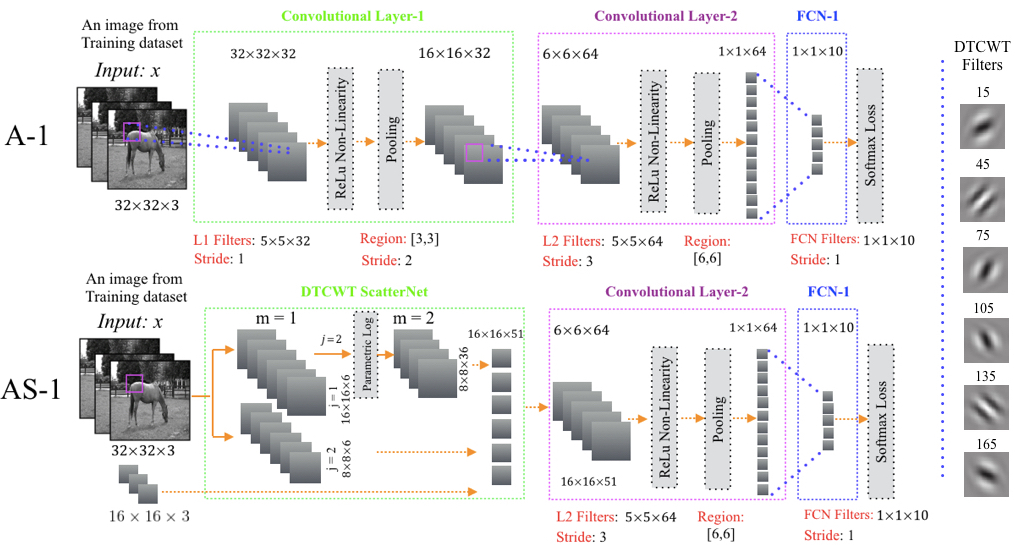}
\caption{{The proposed DTSCNN architecture, termed as AS-1, formed by replacing the first convolutional, ReLu, and pooling layer of A-1 (Table. 1) CNN architecture with the two-layer parametric log based DTCWT ScatterNet. The ScatterNet extracts relatively symmetric translation invariant representations from a multi-resolution image that are processed by the CNN architecture to learn complex representations. However, the illustration shows the feature extraction only for a single image due to space constraints. The invariant information ($U[\lambda_{m = 1}]$) obtained for each R, G and B channel of an image is combined into a single invariant feature by taking an L2 norm of them. Log transformation is applied with parameter $k_{1}$ = 1.1 for scale j = 1. The representations at all the layers ($m = 0  (3)$, $m = 1  (12)$ and $m = 2  (36)$) are concatenated to produce 51*2 (two resolutions) = 102 image representations that are given as input to the mid and back layers of the CNN.}}
\label{fig:scatter00}
\end{figure*}

This paper proposes the DTCWT ScatterNet Convolutional Neural Network (DTSCNN) formed by replacing the first convolutional, relu and pooling layers of the CNN with a two-layer parametric log based DTCWT ScatterNet~\cite{singh}. This extracts relatively symmetric translation invariant representations from a multi-resolution image using the \textit{dual-tree complex wavelet transform} (DTCWT)~\cite{Kingsbury1998} and a parametric log transformation layer. These extracted features, that incorporate edge information similar to the first layers of the networks trained on ImageNet~\cite{zl}, \cite{girshick}, are used by the middle and later CNN layers to learn high-level features. This helps the proposed DTSCNN architecture to converge faster as it has fewer filter weights to learn compared to its corresponding CNN architecture (Section 2). In addition, the CNN layers can learn more complex patterns from the start of learning as it is not necessary to wait for the first layer to learn edges as they are already extracted by the ScatterNet. 

The performance of the DTSCNN architecture is evaluated on (i) Classification error and, (ii) Computational efficiency and the rate of learning with over 50 experiments performed with 14 CNN architectures. The efficient learning of the DSTCNN architectures is demonstrated by their ability to train faster and with lower classification error on small as well as large training datasets, generated from the CIFAR-10 dataset. The DTCWT ScatterNet front-end is also shown to give similar performance to the first convolutional pre-trained layers of CNNs which capture problem specific filter representations, on Caltech-101 as well as CIFAR-10 datasets. A comparison with the state-of-the-art is also presented on both datasets.

  The paper is divided into the following sections. Section 2 presents the parametric log based DTCWT ScatterNet Convolutional Neural Network (DTSCNN). Section 3 presents the experimental results while Section 4 draws conclusions.

\section{Proposed DTSCNN Network}
\label{sec:typestyle}
This section details the proposed DTCWT ScatterNet Convolutional Neural Network (DTSCNN) composed by combining the two-layer parametric log based DTCWT ScatterNet with the later layers (middle and back-end) of the CNN to perform object classification. The ScatterNet (front-end) is first briefly explained followed by the details regarding the (back-end) CNN architecture.
  
The parametric log based DTCWT ScatterNet~\cite{singh} is an improved version (both on classification error and computational efficiency) of the multi-layer Scattering Networks~\cite{Jbruna2013,Oyallon2015,ima,eccv,fergal} that extracts relatively symmetric translation invariant representations from a multi-resolution image using the \textit{dual-tree complex wavelet transform} (DTCWT)~\cite{Kingsbury1998} and parametric log transformation layer. This network extracts feature maps that are denser over scale from multi-resolution images at 1.5 times and twice the size of the input image. Below we present the formulation of the parametric DTCWT ScatterNet for a single input image which may then be applied to each of the multi-resolution images.

\begin{table*}[!t]
\centering
\caption{{Experiments are performed with CNN architectures (derived from LeNet~\cite{LeCun1998}) designed for CIFAR-10 dataset that contain convolutional (CV) layers (L1 to L5) with $b$ number of filters of size $a\times a$, denoted as L-F: $a,b$. The max pooling is performed for a layer within a region of size $c\times c$, denoted as PL-R: [$c,c$]. The network also contains fully connected layers (FCN) that feed the final CNN outputs to a softmax loss function. The architectures are: (i) A-1: 2CV-1FCN (ii) A-2: 3CV-2FCN (iii) A-3: 4CV-3FCN (iv) A-4: 5CV-3FCN.}}

\begin{tabular}{|>{}m{1.2cm}|l| c |c|c |c| c|c |c |c|c|c|c|}
\hline
\multicolumn{1}{c}{Architecture} & \multicolumn{11}{c}{Layers}   \\ 
\hline
\cline{2-9} \hline
& L1-F & PL1-R & L2-F & PL2-R & L3-F & PL3-R & L4-F & L5-F & FCN1 & FCN2 & FCN3 \\
\hline
   & a,b & [c,c] & a,b & [c,c] & a,b & [c,c] & a,b & a,b & a,b & a,b & a,b \\
\cline{2-9} \hline
A-1      & 5,32    & [3,3] & 5,64  & [6,6] & --  & -- & -- & -- & 1,10 & -- & --      \\ \hline
A-2    & 5,32    & [3,3] & 5,32  & [6,6] & 5,64  & [4,4] & -- & -- & 1,32 & 1,10 & --    \\  \hline
A-3 & 5,32    & [3,3] & 5,32  & [3,3] & 5,64  & [3,3] & 4,64 & -- & 1,32 & 1,16 & 1,10    \\ \hline
A-4  & 5,32    & [3,3] & 5,32  & [3,3] & 5,64  & -- & 4,64 & 4,64 & 1,32 & 1,16 & 1,10    \\ \hline
\end{tabular}

\end{table*}

 \begin{table*}[!t]
\centering
\caption{{Parameter values used by the architectures mentioned in Table. 1 for training are: Learning rate = 0.001, Number of Epochs = 300, Weight Decay = 0.0005  and Momentum = 0.9. The batch size is changed according to the number of training samples as mentioned below.
}}
\begin{tabular}{|>{}m{4.7cm}|l| c |c|c  |c| c|c |c |c}
\hline
Training Data Sample Size   & 300 & 500 & 1000 & 2000 & 5000 & 10000 & 25000 & 50000 \\
\cline{2-9} \hline
Batch Size      & 5    & 5 & 10  & 20 & 50 & 100 & 100 & 100     \\ \hline
\end{tabular}
\end{table*}

The invariant features are obtained at the first layer by filtering the input signal $x$ with dual-tree complex wavelets $ \psi_{j,r }$ at different scales ($j$) and six pre-defined orientations ($r$) fixed to $15^\circ, 45^\circ, 75^\circ, 105^\circ, 135^\circ$ and $165^\circ$.  To build a more translation invariant representation, a point-wise $L_{2}$ non-linearity (complex modulus) is applied to the real and imaginary part of the filtered signal:

\begin{equation}
U[\lambda_{m = 1}] = |x\star \psi_{\lambda_{1} }| = \sqrt{|x\star \psi_{\lambda_{1} }^{a}|^2 + |x\star \psi_{\lambda_{1} }^{b}|^2} 
\end{equation}
The parametric log transformation layer is then applied to all the oriented representations extracted at the first scale $j=1$ with a parameter $k_{j=1}$, to reduce the effect of outliers by introducing relative symmetry of pdf, as shown below: 
 
   \begin{equation}
  U1[j] = \log(U[j] + k_{j}), \quad U[j] = |x\star \psi_{j}|, 
\end{equation}
Next, a local average is computed on the envelope $|U1[\lambda_{m = 1}]|$ that aggregates the coefficients to build the desired translation-invariant representation: 

\begin{equation}
S[\lambda_{m = 1}] = |U1[\lambda_{m = 1}]| \star \phi_{2^J}
\end{equation}
The high frequency components lost due to smoothing are retrieved by cascaded wavelet filtering performed at the second layer. The retrieved components are again not translation invariant so invariance is achieved by first applying the L2 non-linearity of eq(2) to obtain the regular envelope:

\begin{equation}
U2[\lambda_{m = 1},\lambda_{m = 2}] = |U1[\lambda_{m = 1}] \star \psi_{\lambda_{m = 2}}|
\end{equation}

and then a local-smoothing operator is applied to the regular envelope ($U2[\lambda_{m = 1},\lambda_{m = 2}]$) to obtain the desired second layer ($m = 2$) coefficients with improved invariance: 

\begin{equation}
S[\lambda_{m = 1},\lambda_{m = 2}] = U2[\lambda_{m = 1},\lambda_{m = 2}] \star \phi_{2^J}
\end{equation}
The scattering coefficients for each layer are:
\vspace{-0.3em}
\begin{equation}
S = \begin{pmatrix}
x \star \phi_{2^J} \\
U1[\lambda_{m = 1}] \star \phi_{2^J}
\\ U2[\lambda_{m = 1},\lambda_{m = 2}] \star \phi_{2^J}
\end{pmatrix}_{j = 2}
\end{equation} 

\begin{table*}[!t]
\centering
\caption{{Classification error (\%) on the CIFAR-10 dataset for the original CNN architectures and their corresponding DTSCNN architectures.}}
\begin{tabular}{|>{}m{4.2cm}|l| c |c|c  |c| c|c |c |c}
\hline
\multicolumn{1}{c}{Architectures} & \multicolumn{8}{c}{Classification Error}   \\ 
\hline
\hline
\multicolumn{1}{c}{Derived from LeNet~\cite{LeCun1998}} & \multicolumn{8}{c}{Training Data Sample Size}   \\ 
\hline
   & 300 & 500 & 1000 & 2000 & 5000 & 10000 & 25000 & 50000 \\
\cline{2-9} \hline
A-1: 2Conv-1FCon      & 77.8    & 73.2 & 70.3  & 66.7 & 61.3  & 54.9 & 45.3 & \cellcolor{gray!50}\textbf{38.1}      \\ \hline
AS-1: DTS-1Conv-1FCon   & \cellcolor{gray!50}\textbf{69.4}   & \cellcolor{gray!50}\textbf{65.8} & \cellcolor{gray!50}\textbf{60.1} & \cellcolor{gray!50}\textbf{58.9} & \cellcolor{gray!50}\textbf{52.7}  & \cellcolor{gray!50}\textbf{54.7} & \cellcolor{gray!50}\textbf{40.4} & 38.7      \\ 
\hline
A-2: 3Conv-2FCon       & 66.8      & 62.0 & 57.5 & 52.8 & 46.1 & 40.1 & \cellcolor{gray!50}\textbf{32.7} & \cellcolor{gray!50}\textbf{27.3} \\  \hline
AS-2: DTS-2Conv-2FCon & \cellcolor{gray!50}\textbf{63.7} & \cellcolor{gray!50}\textbf{55.1} & \cellcolor{gray!50}\textbf{49.5}  & \cellcolor{gray!50}\textbf{43.7}  & \cellcolor{gray!50}\textbf{39.1} & \cellcolor{gray!50}\textbf{40.0} & 33.8 & 28.3     \\ \hline
A-3: 4Conv-3FCon      & 62.2    & 57.4 & 51.0 & 46.8 &  40.1 & 35.1 & 29.2 & 24.2 \\ \hline
AS-3:DTS-3Conv-3FCon       & \cellcolor{gray!50}\textbf{56.8}     & \cellcolor{gray!50}\textbf{54.9} & \cellcolor{gray!50}\textbf{50.9} & \cellcolor{gray!50}\textbf{45.3} & \cellcolor{gray!50}\textbf{39.7} & \cellcolor{gray!50}\textbf{34.9} & \cellcolor{gray!50}\textbf{28.7} & \cellcolor{gray!50}\textbf{24.1}      \\ \hline
A-4: 5Conv-3FCon      & \cellcolor{gray!50}\textbf{58.4}    & 54.4  & 47.4 & 41.8  & \cellcolor{gray!50}\textbf{35.0} & 32.2 & 25.7 & 22.1\\
\hline
AS-4: DTS-4Conv-3FCon      & 59.8   &\cellcolor{gray!50}\textbf{54.0} & \cellcolor{gray!50}\textbf{47.3} & \cellcolor{gray!50}\textbf{41.3} & 38.4 & \cellcolor{gray!50}\textbf{31.8} & \cellcolor{gray!50}\textbf{25.2} & \cellcolor{gray!50}\textbf{22.0}     \\ 
\hline
\multicolumn{1}{c}{Standard Deep Architectures} & \multicolumn{8}{c}{Training Data Sample Size}   \\
\hline 
A-5: NIN~\cite{NIN}   & 89.2 & 84.4   & 45.5 & 34.9 & 27.1 & 18.8 & \cellcolor{gray!50}\textbf{13.3} & \cellcolor{gray!50}\textbf{8.1}     \\ 
\hline
AS-5: DTS-NIN &  \cellcolor{gray!50}\textbf{83.2} & \cellcolor{gray!50}\textbf{80.1} & \cellcolor{gray!50}\textbf{41.0} & \cellcolor{gray!50}\textbf{32.2} & \cellcolor{gray!50}\textbf{25.3} & \cellcolor{gray!50}\textbf{18.4} & 13.4 & 8.2    \\ 
\hline
A-6: VGG~\cite{vgg}     & 89.9   & 89.7 & 89.1 & 59.6 & 36.6 & 28 & 16.9 & \cellcolor{gray!50}\textbf{7.5}     \\ 
\hline
AS-6: DTS-VGG     & \cellcolor{gray!50}\textbf{83.5}   & \cellcolor{gray!50}\textbf{82.8} & \cellcolor{gray!50}\textbf{81.6} & \cellcolor{gray!50}\textbf{56.7} & \cellcolor{gray!50}\textbf{34.9} & \cellcolor{gray!50}\textbf{27.2} & \cellcolor{gray!50}\textbf{16.9} & 7.6     \\ 
\hline
A-7: WResNet~\cite{wresnet}      &  87.2 & 53.2 & 43.2 & 31.1 & 18.8 & 13.6 & 10.1 & 3.6   \\ 
\hline
AS-7: DTS-WResNet  & \cellcolor{gray!50}\textbf{81.2}   & \cellcolor{gray!50}\textbf{49.8} & \cellcolor{gray!50}\textbf{41.2} & \cellcolor{gray!50}\textbf{30.1} & \cellcolor{gray!50}\textbf{18.6} & \cellcolor{gray!50}\textbf{13.5} & \cellcolor{gray!50}\textbf{9.9} & \cellcolor{gray!50}\textbf{3.6}  \\ 
\hline
\end{tabular}
\end{table*}

Next, the proposed DTSCNN architectures (AS1 to AS4) are realized by replacing the first convolutional layer of the A-1 to A-4 CNN architectures with the ScatterNet (described above), as shown in Fig. 1. The four CNN architectures (A-1 to A-4, shown in Table. 1) are derived from the LeNet~\cite{LeCun1998} architecture because they are relatively easy to train due to its small memory footprint. In addition to the derived architectures, the DTSCNN is also realized by using ScatterNet as the front-end of three standard deep architectures namely; Network in Network (NIN)~\cite{NIN} (A-5), VGG~\cite{vgg} (A-6), and wide ResNet~\cite{wresnet} (WResNet) (A-7). The DTSCNN architectures (AS-5, AS-6, AS-7) for the standard architectures (NIN (A-5), VGG (A-6), WResNet (A-7)) are again obtained by removing the first convolutional layer of each network and replacing it with the ScatterNet. The architectures are trained in an end-to-end manner by Stochastic Gradient Descent with softmax loss until convergence. 

\section{Experimental Results}
The performance of the DTSCNN architecture is demonstrated on CIFAR-10 and Caltech-101 datasets with over 50 experiments performed with 14 CNN architectures of increasing depth on (i) Classification error and, (ii) Computational efficiency and the rate of learning. The generic nature of the features extracted by the DTCWT ScatterNet is shown by an equivalent performance to the pre-trained CNN front-ends. The details of the datasets and the results are presented below. 

\subsection{Datasets}
The CIFAR-10~\cite{cifar} dataset contains a total of 50000 training and 10000 test images of size $32\times32$. The efficient learning of the proposed DTSCNN network is measured on 8 training datasets of sizes: 300, 500, 1000, 2000, 5000, 10000, 25000 and 50000 images generated randomly by selecting the required number of images from the full 50000 training dataset. It is made sure that an equal number of images per object class are sampled from the full training dataset. For example, a training dataset sample size of 300 will include 30 images per class. The full test set of 10000 images is used for all the experiments. 

Caltech-101~\cite{caltech} dataset contains 9K images each of size 224$\times$224 labeled into 101 object categories and a background class. The classification error on this dataset is measured on 3 randomly generated splits of training and test data, so that each split contains 30 training images per class, and up to 50 test images per class. In each split, 20\% of training images were used as a validation set for hyper-parameter selection. Transfer learning is used to initialize the filter weights for the networks that are used to classify this dataset because the number of training samples are not sufficient to train the networks from a random start.

\subsection{Evaluation and Comparison on Classification Error}
The classification error is recorded for the proposed DTSCNN architectures and compared with the derived (A-1 to A-4) as well as the standard (A-5 to A-7) CNN architectures, for different training sample sizes, as shown in Table. 1. The parameters used to train the CNN architectures are shown in Table. 2. The classification error corresponds to the average error computed for 5 repetitions. 

It can be observed from Table. 3 that the difference in classification error for the derived CNN architectures (A-1 to A-4) is at around 9\% for the small training datasets with $\leq$ 1000 training images. This difference in error reduces with the increase in the size of the training dataset and with increase in the depth of the architectures as shown in Table. 3. In fact, the deeper CNN architectures such 5CV-3FCN (A-4) outperformed their corresponding DTSCNN architectures by a small margin. 

\begin{figure*}[!t]
\begin{subfigure}{.5\textwidth}
  \centering
  \includegraphics[width=0.92\linewidth,height = 4.8cm]{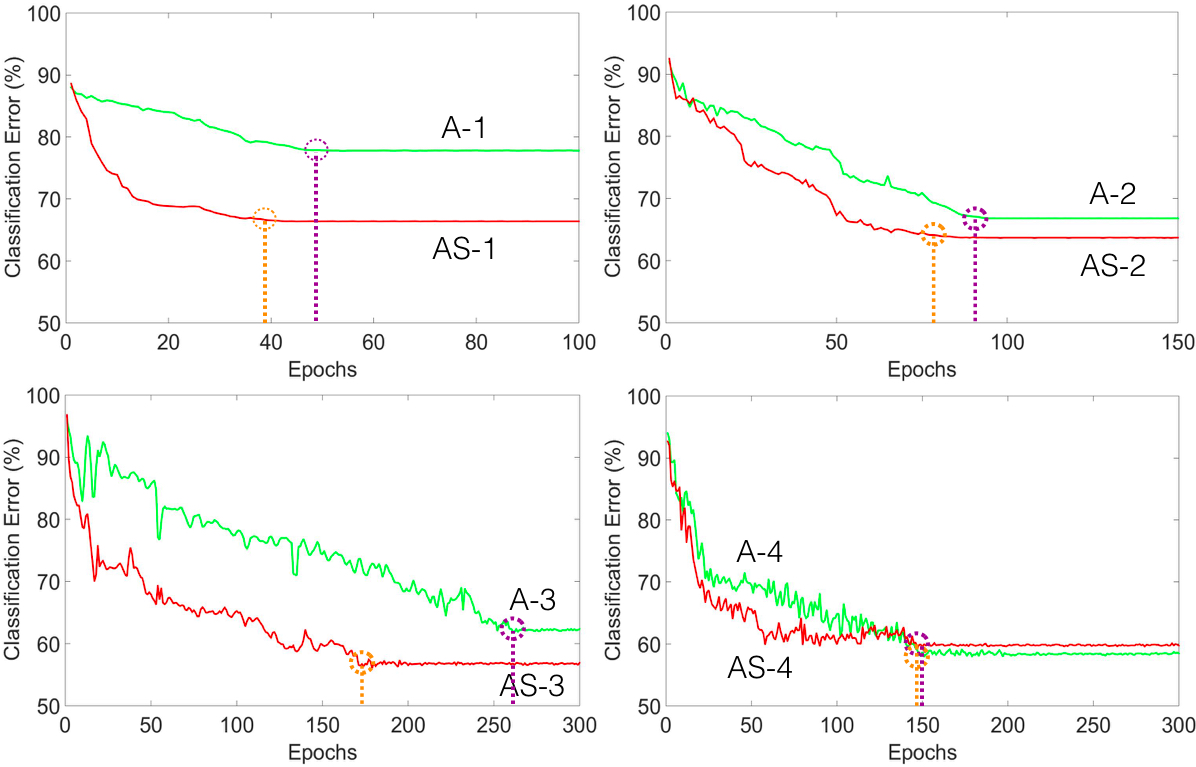}
  \caption{Training dataset sample size: 300}
  \label{fig:sfig1}
\end{subfigure}
\begin{subfigure}{.5\textwidth}
  \centering
  \includegraphics[width=0.92\linewidth]{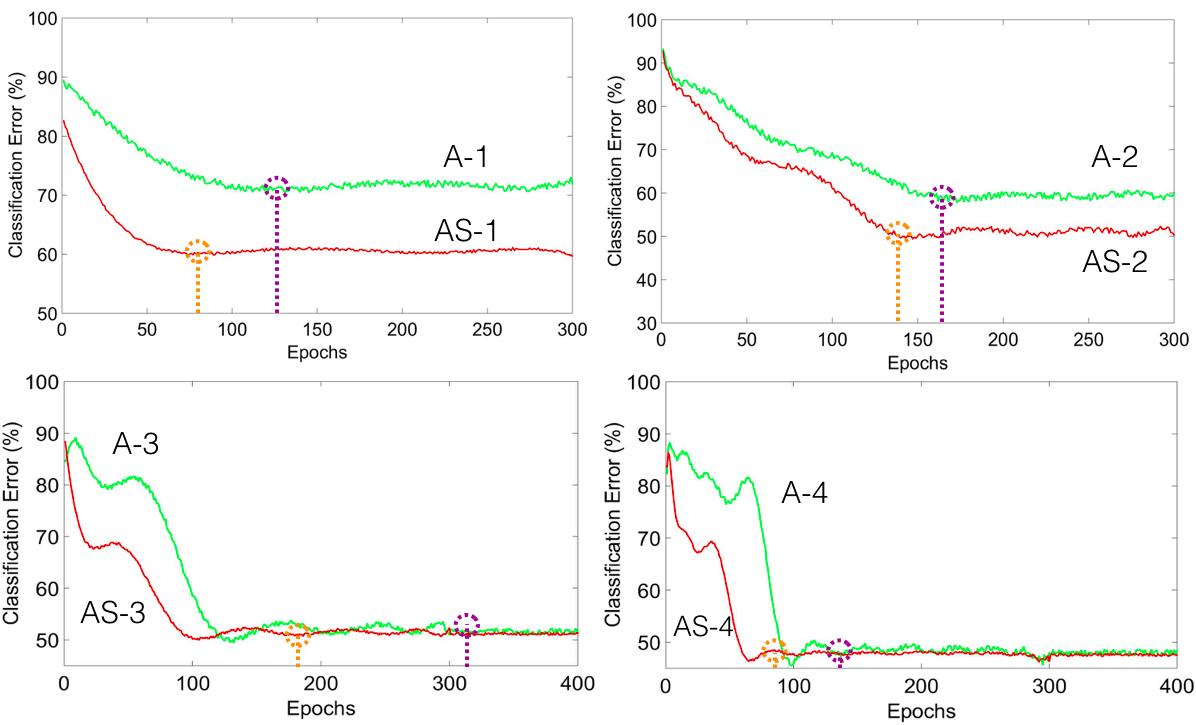}
  \caption{Training dataset sample size: 1000}
  \label{fig:sfig2}
\end{subfigure}

\begin{subfigure}{.5\textwidth}
  \centering
  \includegraphics[width=0.92\linewidth]{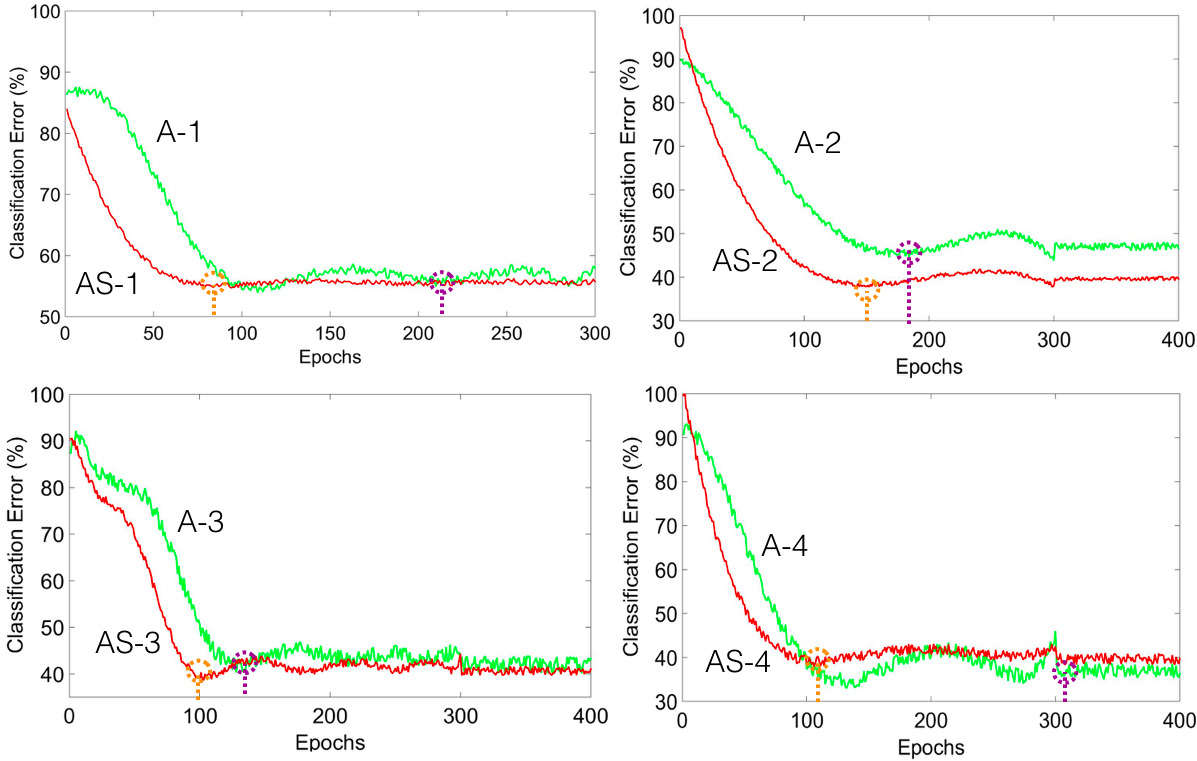}
  \caption{Training dataset sample size: 10000}
  \label{fig:sfig1}
\end{subfigure}
\begin{subfigure}{.5\textwidth}
  \centering
  \includegraphics[width=0.94\linewidth]{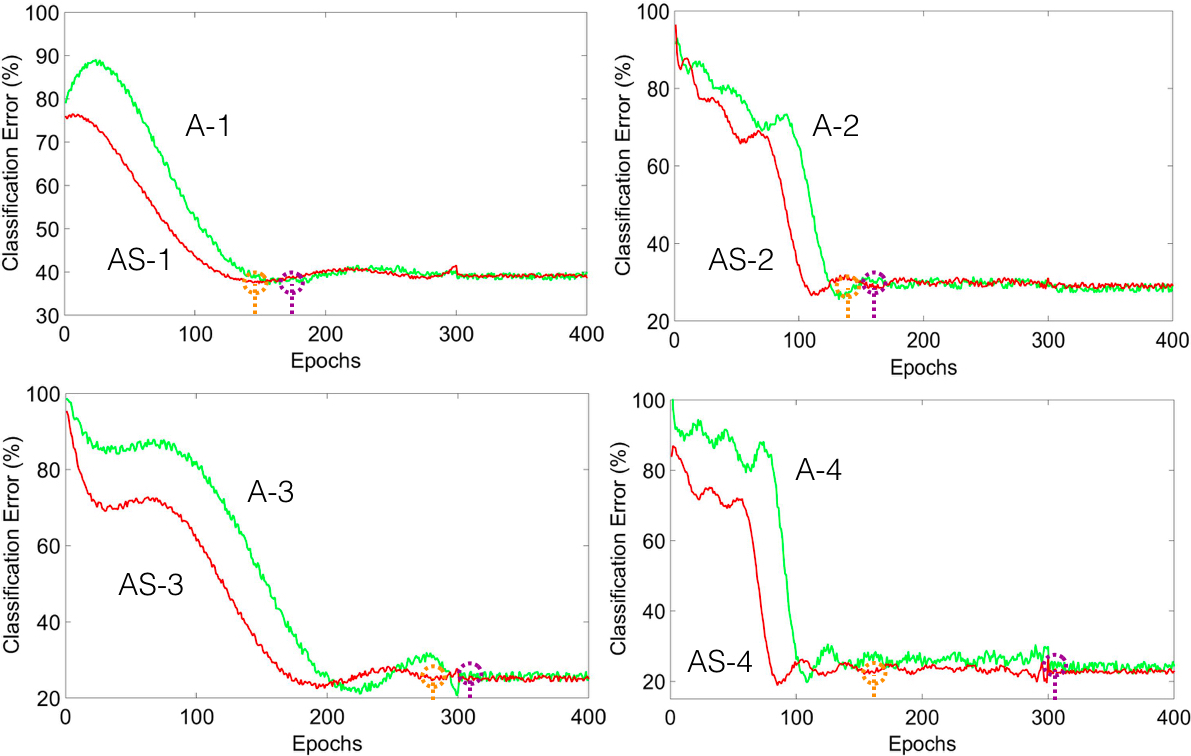}
  \caption{Training dataset sample size: 50000}
  \label{fig:sfig2}
\end{subfigure}
\begin{subfigure}{.5\textwidth}
  \centering
  \includegraphics[width=0.90\linewidth]{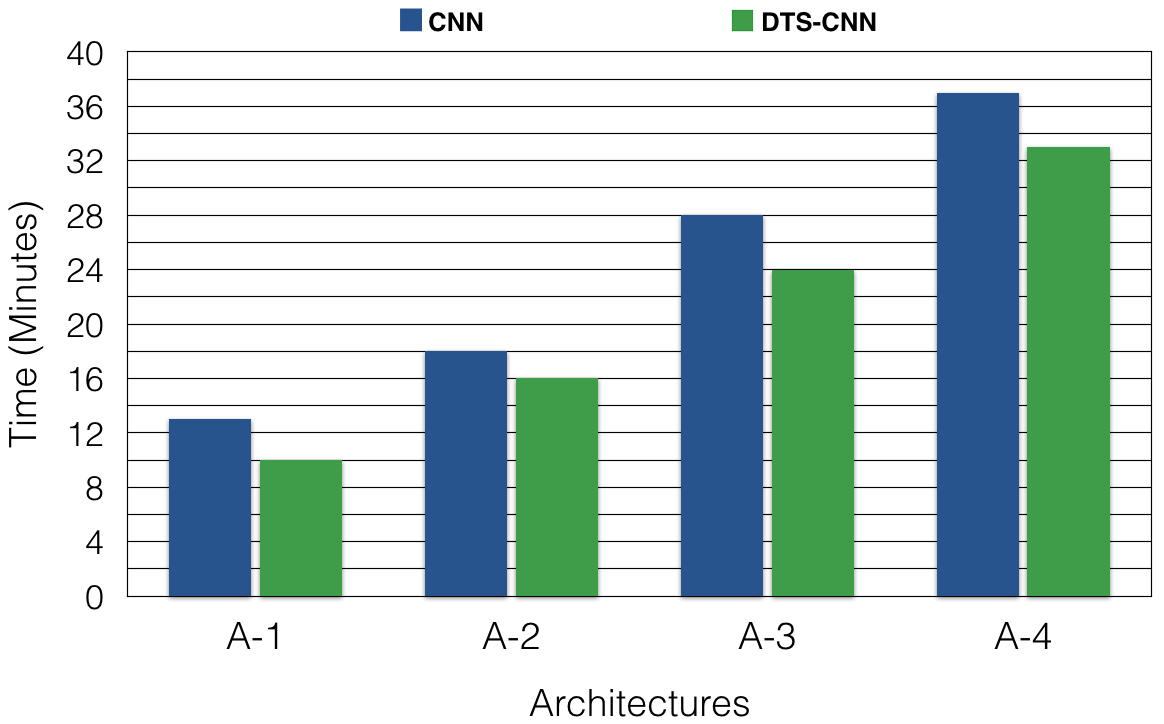}
  \caption{Computational time for convergence for 5000 training size}
  \label{fig:sfig2}
\end{subfigure}
\begin{subfigure}{.5\textwidth}
  \centering
  \includegraphics[width=0.90\linewidth]{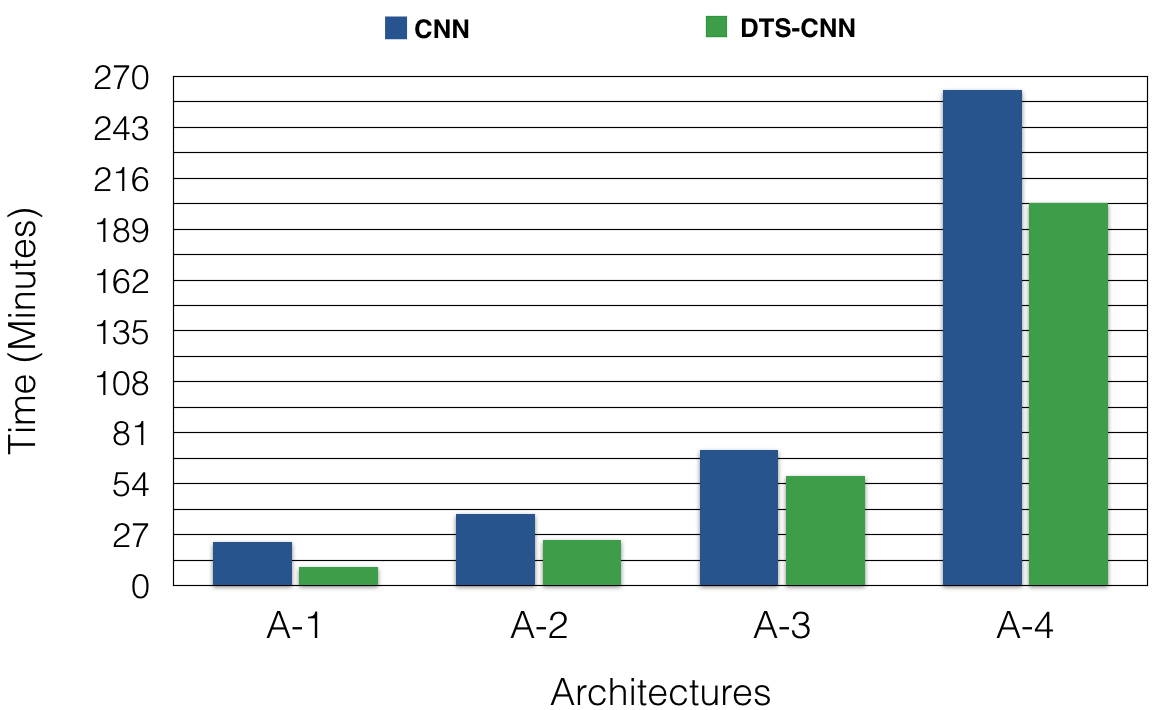}
  \caption{Computational time for convergence for 50000 training size}
  \label{fig:sfig2}
\end{subfigure}
\caption{{Graphs show the faster convergence and rate of learning of the DTSCNN derived architectures (AS-1 to AS-4) compared to the CNN (A-1 to A-4) architectures for a range of small and large training data sizes. An architecture is considered to have converged at a specific epoch when the error value for the subsequent epochs changes within 2\% of the error value at that specific epoch. The convergence is marked on the epoch axis using an orange dotted line for the DTSCNN architecture and a purple dotted line for the CNN architectures. The orange line has a lower epochs value as compared to the purple line indicating the faster convergence. Computational time for convergence for the original CNN and the corresponding DSTCNN networks measured to within 2\% of the final converged error value is also shown for a small (5000) and large (50000) training dataset.}}
\label{fig:fig}
\end{figure*}

A similar trend in classification error is also observed for the standard more deeper CNN architectures (A-5 to A-7). The difference in classification error is large between the DTSCNN (AS-5 to AS-7) and the original (A-5 to A-7), architectures for small training datasets while both class of architectures produce a similar classification error for datasets with large training size. In fact, the wide ResNet (WResNet) (A-7) and its corresponding DTSCNN architecture (AS-7) result in the same classification error of 3.6\%. 

\begin{figure*}[!t]
\begin{subfigure}{.5\textwidth}
  \centering
  \includegraphics[width=0.93\linewidth,height = 2.8cm]{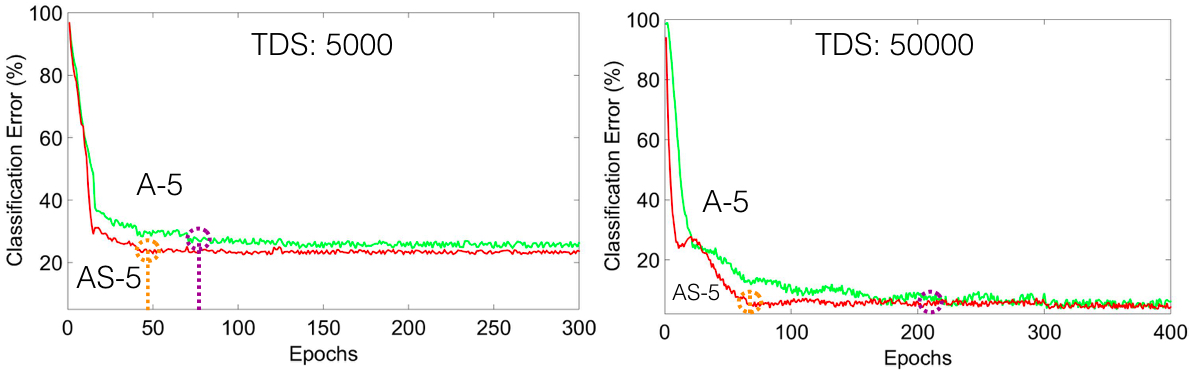}
  \caption{Network in Network (NIN)}
  \label{fig:sfig1}
\end{subfigure}
\begin{subfigure}{.5\textwidth}
  \centering
  \includegraphics[width=0.93\linewidth,,height = 2.8cm]{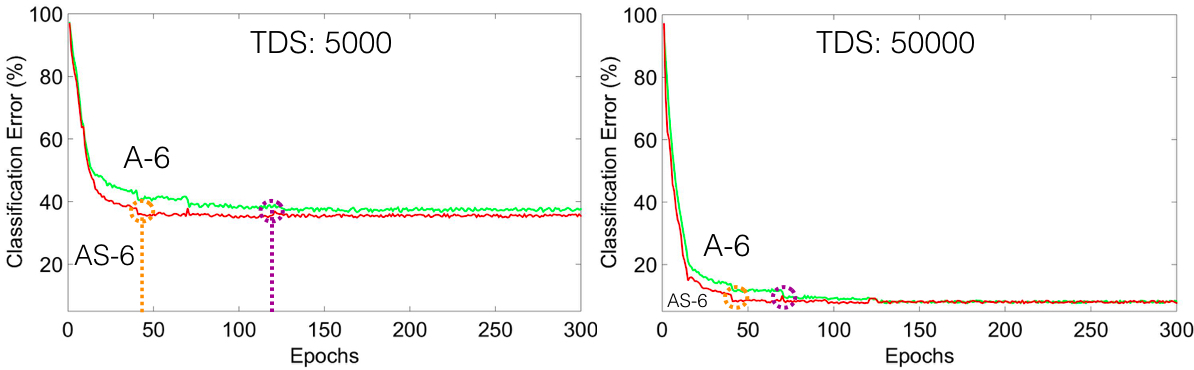}
  \caption{Visual Geometry Group (VGG) Network}
  \label{fig:sfig2}
\end{subfigure}

\begin{subfigure}{0.5\textwidth}
  \centering
  \includegraphics[width=0.93\linewidth,,height = 2.9cm]{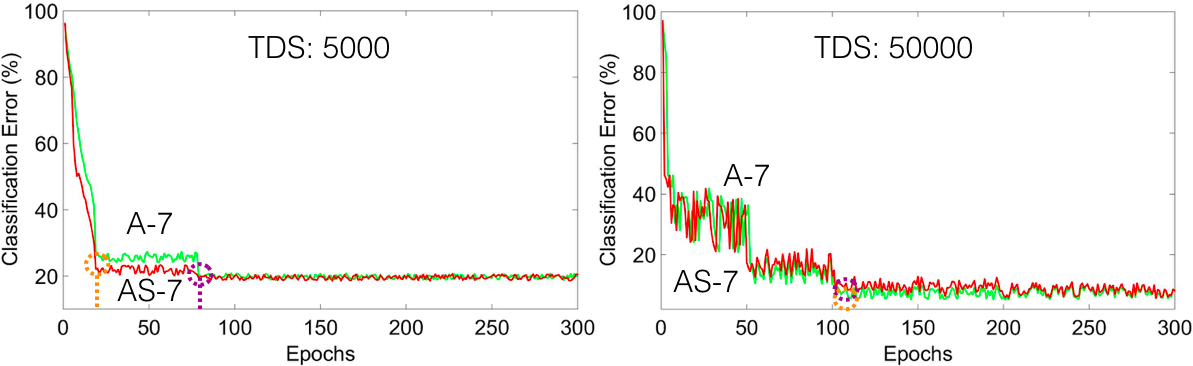}
  \caption{Wide Residual Network (WResNet)}
  \label{fig:sfig1}
\end{subfigure}
\begin{subfigure}{0.5\textwidth}
  \centering
  \includegraphics[width=0.76\linewidth,,height = 3.2cm]{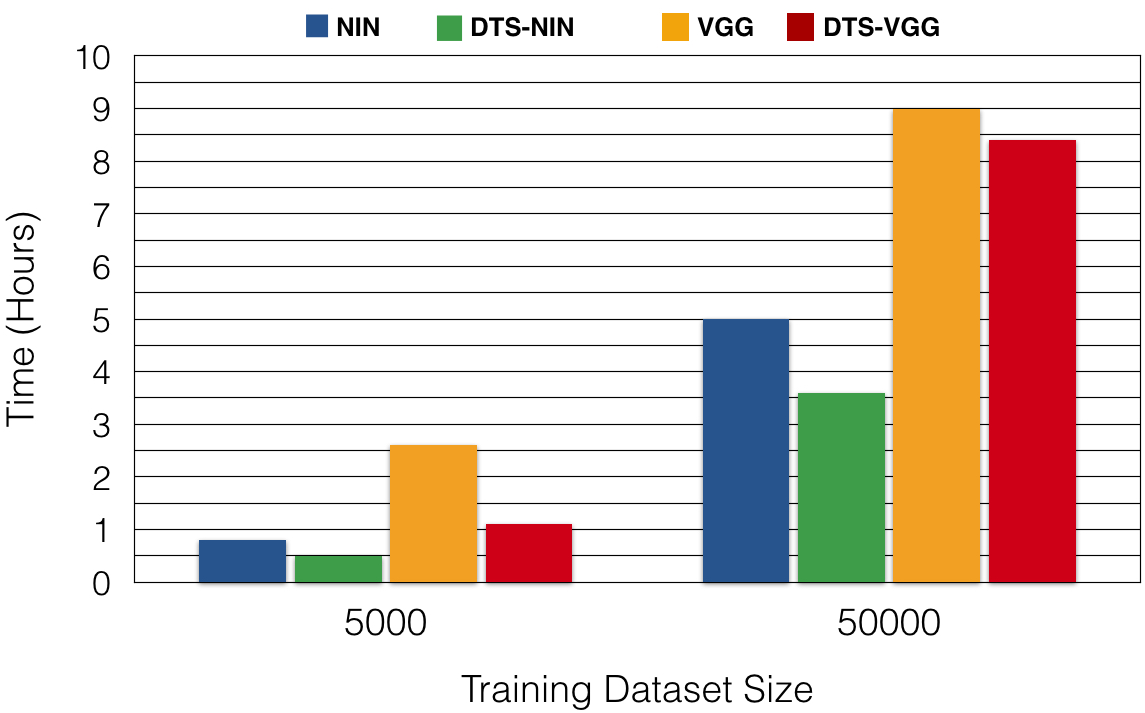}
  \caption{Computational time for convergence.}
  \label{fig:sfig1}
\end{subfigure}

\caption{{Graphs show the faster convergence and rate of learning of the DTSCNN standard deep architectures (AS-5 to AS-7) compared to the CNN (A-5 to A-7) architectures for a small (5000) and large (50000) training dataset. An architecture is considered to have converged at a specific epoch when the error value for the subsequent epochs changes within 2\% of the error value at that specific epoch. The convergence is marked on the epoch axis using an orange dotted line for the DTSCNN architecture and a purple dotted line for the CNN architectures. The orange line has a lower epochs value as compared to the purple line indicating the faster convergence. Computational time for convergence (hours) for NIN (A-5) and VGG (A-6) standard deep architectures and corresponding DTSCNN architectures for a small (5000) and a large (50000) training dataset is also presented.}}
\label{fig:fig}
\end{figure*}

\subsection{Analysis on Computational Efficiency and Learning}
This section compares the efficient and faster learning of the proposed DTSCNN architectures against the CNN architectures (A-1 to A-4) derived from LeNet~\cite{LeCun1998} as well as the standard (A-5 to A-7) deep learning architectures for a range of small and large training dataset sizes. 

The DTSCNN architectures have a higher rate of learning or faster converge than the original CNN architectures because the numbers of filter weights required to be learned are smaller but also because the ScatterNet extracts edge representations that allow the later CNN layers to learn high-level features from the first epoch onwards. The faster convergence is shown for both the derived (A-1 to A-4) and standard deep architectures (A-5 to A-7) as shown in Fig. 2 and Fig. 3, respectively. An architecture is considered to have converged at a specific epoch when the error value for the subsequent epochs changes within 2\% of the error value at that specific epoch. The convergence is marked on the epoch axis using an orange dotted line for the DTSCNN architecture and a purple dotted line for the CNN architectures. As observed from Fig. 2 and Fig. 3, the orange line has a lower epoch value as compared to the purple line indicating the faster convergence. 

The time required for training the original and their corresponding DSTCNN architectures is presented for a small (5000) and large training dataset (50000) for both the derived (A-1 to A-4) and standard deep architectures (A-5 to A-6), as shown in Fig. 2 and Fig. 3, respectively. The time for convergence is again measured to within 2\% of the final converged error value. As observed from both figures, the training time is higher for the original networks than the DTSCNN networks because of the reasons mentioned above.

The networks are trained using the MatConvNet~\cite{mnet} package on a server with a NVIDIA GeForce 7800 GTX card.

\subsection{Comparison with Pre-trained CNN First Layers}
The classification performance of the DTCWT ScatterNet front-end is compared with the first pre-trained convolutional layer for the Network in Network (NIN)~\cite{NIN} and the Visual Geometry Group convolutional (VGG)~\cite{vgg} architectures, on Caltech-101 and CIFAR-10 datasets. The filter weights for both the NIN and the VGG networks are initialized with the weights obtained from their models pre-trained on ImageNet (found here~\cite{bvlc}). The first layers for both the architectures are fixed to be the ScatterNet and the pre-trained convolutional layer, while the filter weights only in later layers are fine-tuned using the training images of CIFAR-10 and Caltech-101. The ScatterNet front-end gives similar performance to the pre-trained first convolutional layer on classification error for both datasets as shown in Table. 4. For this experiment, dropout, batch normalization and data augmentation with crops and horizontal flips were utilized~\cite{torch}. The use of the NIN network is preferred as it gives similar performance to the VGG network while being 4 times faster.

\begin{table}[!h]%
\centering
        \caption{Table shows the comparison on classification error (\%) between the DTCWT ScatterNet (DTS) front-end and the first convolutional layer pre-trained on ImageNet for NIN~\cite{NIN} and VGG~\cite{vgg} architectures for Caltech-101 and CIFAR-10 datasets.T-NIN: Transfer-NIN, T-VGG: Transfer-VGG}
            \begin{tabular}{|>{}m{1.55cm}|c| c |c|c|c|}
\hline
\multicolumn{1}{c}{Dataset} & \multicolumn{4}{c}{State-of-the-art Architectures}   \\ 
\hline
    & T-NIN & DTS-NIN  & T-VGG& DTS-VGG\\
\cline{2-4} \hline
\small{Caltech-101}   & 12.3 & \cellcolor{gray!50}\textbf{12.26}  & \cellcolor{gray!50}\textbf{8.78} & 9.23 \\ 
\small{CIFAR-10}   & \cellcolor{gray!50}\textbf{8.25} & 8.34   & \cellcolor{gray!50}\textbf{8.31} & 9.02 \\ 
\hline
\end{tabular}
\end{table}

\subsection{Comparison with the state-of-the-art}
This section compares the architectures that produced the best classification performance with the state-of-the-art on CIFAR-10 and Caltech-101. The DTS-WResNet (AS-7) and DTS-VGG (AS-6) resulted in the best classification performance on CIFAR-10 and Caltech-101 with 3.6\% and 8.08\% classification error, respectively.

The DTS-WResNet (AS-7) architecture is compared with the state-of-the-art CNN architectures on CIFAR-10. DTS-WResNet outperformed these architectures as shown in Table 5.

\begin{table}[!h]%
\centering

        \caption{Table shows the comparison on classification error (\%) between the DTCWT ScatterNet ResNet (DTS-WResNet) Architecture with the state of the art architectures on the CIFAR-10 dataset. DW: DTS-WResNet, NIN: Network in Network~\cite{NIN}, VGG~\cite{vgg}, DSN: Deeply Supervised Networks~\cite{dsn}, MON: Max-Out Networks~\cite{mon}, E-CNN: Exemplar CNN~\cite{ecnn}}
            \begin{tabular}{>{}m{1.20cm}|c| c ccccc}
\hline
\multicolumn{1}{c}{Dataset} & \multicolumn{6}{c}{State-of-the-art Architectures}   \\ 
\hline
    & DW  & VGG & E-CNN & NIN & DSN & MON \\
\cline{2-4} \hline
Cifar-10 & \cellcolor{gray!50}\textbf{3.6} & 7.5 & 8.0 & 8.1 & 8.2 & 9.3\\ 
\hline
\end{tabular}
\end{table}

Next, the DTS-VGG (AS-6) architecture is compared against the state-of-the-art CNN architectures for the Caltech-101 dataset. On this dataset, the DTS-VGG outperformed some of the architectures while produced a marginally lower classification performance for others (Table 6).

\begin{table}[!h]%
\centering

        \caption{Table shows the comparison on classification error (\%) between the DTCWT ScatterNet VGG (DTS-VGG) Architecture with the state of the art architectures on the Caltech-101 dataset. DTS-VGG: DV, SPP: Spatial Pyramid Pooling~\cite{he}, VGG~\cite{vgg}, E-CNN: Exemplar CNN~\cite{ecnn}, EP: Epitomic Networks~\cite{epi}, ZF: Zieler and Fergus~\cite{zf}}
            \begin{tabular}{>{}m{1.60cm}|c| c ccccc}
\hline
\multicolumn{1}{c}{Dataset} & \multicolumn{6}{c}{State-of-the-art Architectures}   \\ 
\hline
    & DV  & SPP & VGG & E-CNN & EP & ZF\\
\cline{2-4} \hline
\small{Caltech-101} & {8.78} & \cellcolor{gray!50}\textbf{6.6} & 7.3 & 8.5 & 12.2 & 13.5 \\ 
\hline
\end{tabular}
\end{table}

\section{Conclusion}
The proposed DTSCNN architectures, when trained from scratch, outperforms the corresponding original CNN architectures on small datasets by a useful margin. For larger training datasets, the proposed networks give similar error compared to the original architectures. Faster rate of convergence is observed in both cases for shallow as well as deep architectures.

The DTCWT scattering front-end is mathematically designed to deal with all edge orientations equally and with 2 or more scales, as required. The generic nature of the DTCWT scattering front-end is shown by its similar classification performance to the front-end of learned networks, on two different datasets. The generic features are likely to give it wide applicability to both small and large image datasets as it gives lower (small dataset) or similar classification error (large dataset), with faster rates of convergence. 

Future work includes extending the DTCWT Scattering Network front-end for other learning frameworks, with a view to improving learning rates further. 

{
\bibliographystyle{ieee}
\bibliography{egbib}
}

\end{document}